\documentclass{article}
\usepackage{spconf,amsmath,graphicx}
\usepackage{amsmath}
\usepackage{amsfonts}   
\usepackage{multirow}\usepackage{url} 

\title{U-Net with Hadamard Transform and DCT Latent Spaces for Next-day Wildfire Spread Prediction}
\name{Yingyi Luo$^{1}$, Shuaiang Rong$^{1}$, Adam Watts$^{2}$, Ahmet Enis Cetin$^{1}$\thanks{USDA Directorate of Forestry and National Science Foundation Grants IDEAL 2217023, and 2531376.)}}
\address{
  $^{1}$University of Illinois Chicago, Department of Electrical and Computer Engineering, Chicago, IL, USA \\
  $^{2}$US Forest Service, Pacific Wildland Fire Science Laboratory, Seattle, WA, USA
}
\begin{document}
%
\maketitle
\begin{abstract}
We developed a lightweight and computationally efficient tool for next-day wildfire spread prediction using multi-modal satellite data as input. The deep learning model, which we call Transform Domain Fusion UNet (TD-FusionUNet), incorporates trainable Hadamard Transform and Discrete Cosine Transform layers that apply two-dimensional transforms, enabling the network to capture essential "frequency" components in orthogonalized latent spaces.
Additionally, we introduce custom preprocessing techniques, including random margin cropping and a Gaussian mixture model, to enrich the representation of the sparse pre-fire masks and enhance the model's generalization capability. The TD-FusionUNet is evaluated on two datasets which are the Next-Day Wildfire Spread dataset released by Google Research in 2023, and WildfireSpreadTS dataset.
Our proposed TD-FusionUNet achieves an F1 score of 0.591 with 370k parameters, outperforming the UNet baseline using ResNet18 as the encoder reported in the WildfireSpreadTS dataset while using substantially fewer parameters. These results show that the proposed latent space fusion model balances accuracy and efficiency under a lightweight setting, making it suitable for real time wildfire prediction applications in resource limited environments.

\end{abstract}
\section{Introduction}
\label{sec:intro}

Wildfires are among the most devastating natural hazards, causing great economic losses and casualties. In 2024, the United States experienced more than 64,000 wildfires, which burned over 8.9 million acres nationwide \cite{nicc2024}. In January 2025, a series of catastrophic wildfires in Southern California, fueled by drought, burned over 50,000 acres, destroyed more than 16,000 structures, and resulted in at least 29 deaths \cite{calfire2025incidents}. The broader economic loss has also increased rapidly. It is reported that wildfires caused over \$81.6 billion in damages from 2017 to 2021, nearly ten times the \$8.6 billion recorded between 2012 and 2016 \cite{iglesias2022fires}. 

With the frequency and severity of wildfires continuing to rise due to climate change, there is an urgent need for novel prediction methods for effective mitigation and disaster response \cite{ostoja2023nca}\cite{gunay2012entropy}\cite{pan2022deep}. In particular, next-day fire spread prediction, which estimates the areas likely to burn, provides critical information for the optimal allocation of resources and rapid emergency response. Traditional approaches of wildfire spread prediction rely on simulators ranging from empirical to deterministic models \cite{singh2025review}, using statistical analysis of historical data (e.g., regression, correlation) \cite{singh2021land}\cite{duff2021wildfire}\cite{eden2020fire} and dynamic tools that integrate physical models with numerical simulations to represent fire behavior under varying climatic conditions \cite{rahman2018forest}. However, these methods often struggle to capture the complex, nonlinear interactions among environmental variables. The rapid advancement of deep learning provides opportunities for data-driven approaches capable of learning spatiotemporal patterns directly from multi-modal observations, which are critical for improving predictive accuracy \cite{digiuseppe2025fire}. 

Earlier studies \cite{hodges2019wildland}\cite{kantarcioglu2023fire} applied two-layer convolutional neural networks (CNNs) to predict fire occurrence from synthetic datasets or GIS-based remote sensing data. Deep learning neural networks for wildfire prediction face persistent challenges, including high computational cost and limited data availability \cite{Radke2019FireCastLD}. While recent studies \cite{huot2022wildfire}\cite{shadrin2024wildfire} have employed widely used deep learning models such as CNN autoencoders \cite{dong2017learningdeeprepresentationsusing} and UNet/UNet++ \cite{gerard2023wildfirespreadts}\cite{ronneberger2015unetconvolutionalnetworksbiomedical}\cite{zhou2018unetnestedunetarchitecture} for segmentation-based fire spread prediction, most efforts have emphasized data collection and dataset development rather than identifying an efficient deep neural network structure.

\begin{figure*}[t]
  \centering
  \includegraphics[width=\textwidth]{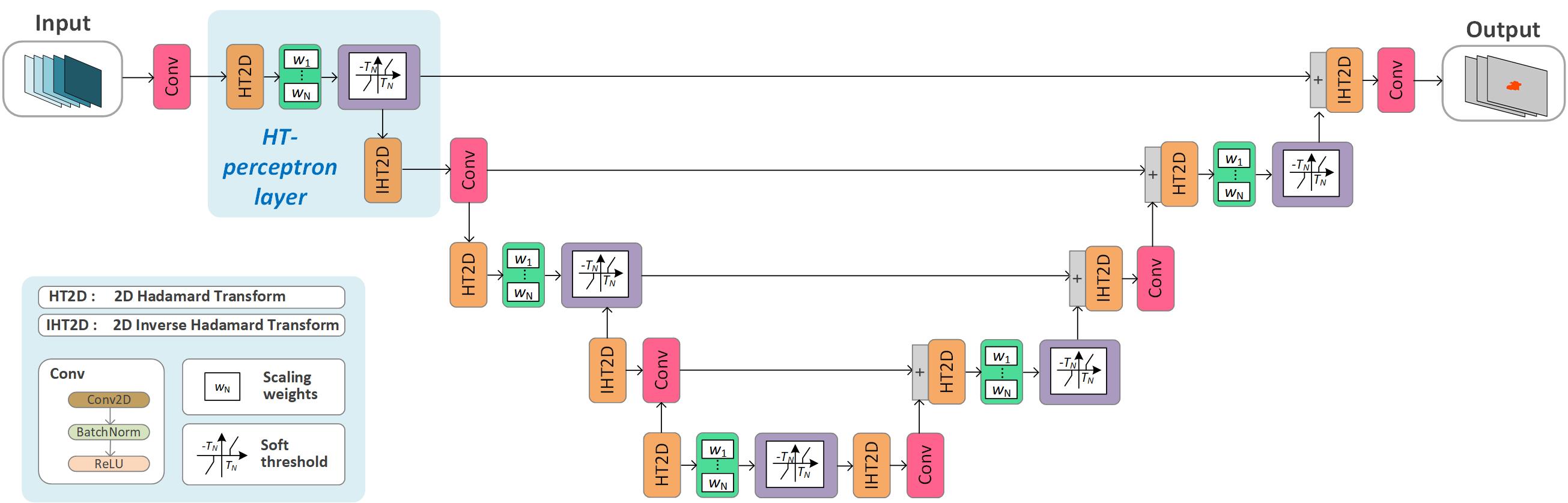}
  \caption{HT-Unet structure}
  \label{fig1}
\end{figure*}
\begin{figure*}[t]
  \centering
  \includegraphics[width=\textwidth]{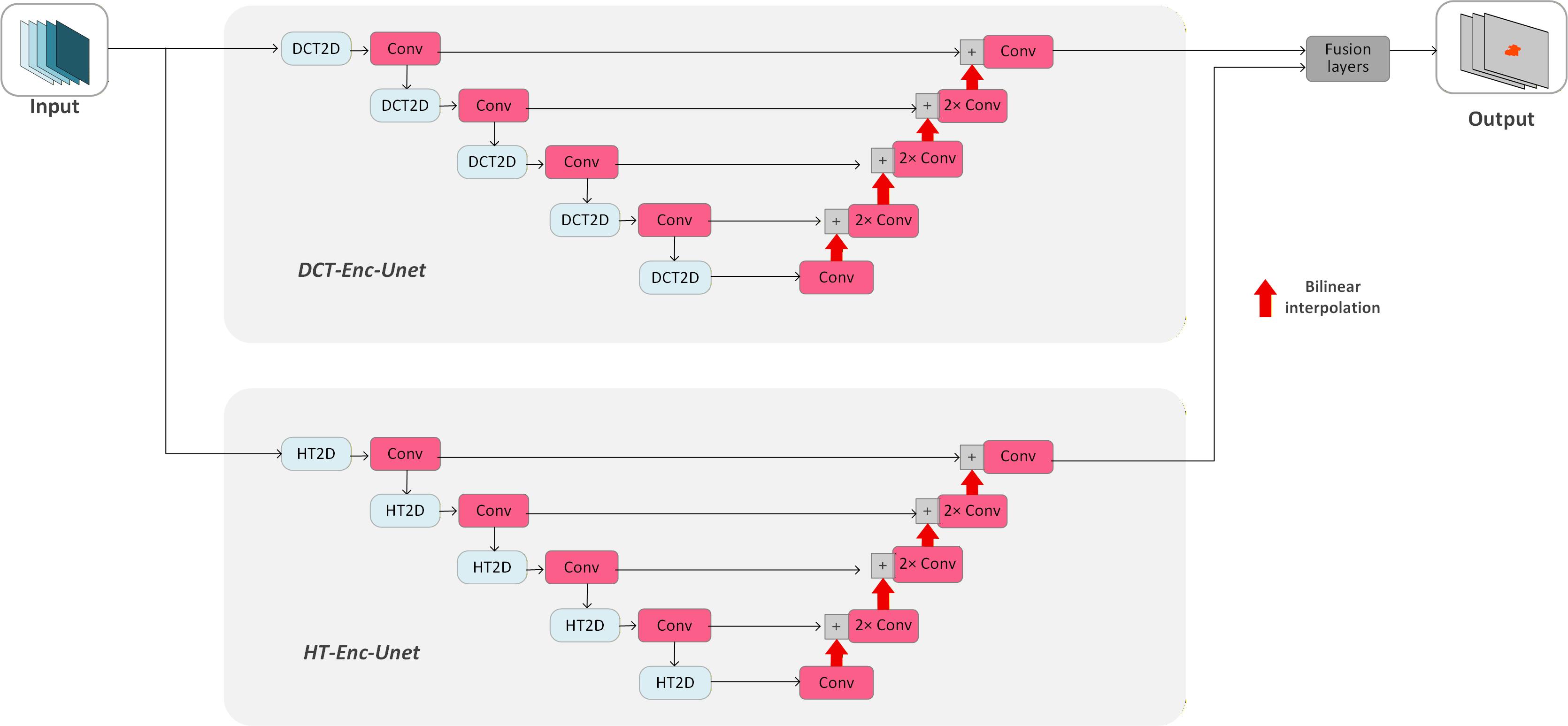}
  \caption{TD-FusionUNet structure}
  \label{DCTHT}
\end{figure*}
To address these issues, we propose a lightweight and efficient Transform domain UNet (TD-FusionUNet) that leverages orthogonal transform-domain feature learning to improve prediction accuracy while maintaining low computational complexity.
%
The TD-FusionUNet improves the traditional U-Net architecture by incorporating trainable Hadamard transform and DCT layers, which applies a 2D Hadamard Transform, and 2D DCT followed by learnable scaling and soft-thresholding in the transform domain, before reconstructing the features with inverse transforms. We could have also used the Fourier transform but it requires complex arithmetic therefore we selected the Hadamard and DCT to establish orthogonal latent spaces.
%
%
In addition, we develop custom preprocessing techniques to address data imbalance and sparsity.
Random margin cropping is applied to balance the representation of burned and unburned pixels, enriching the binary dataset and improving the model’s ability to capture fire spread boundaries.
In addition, a mixed Gaussian softening model is employed to generate synthetic samples, thereby complementing missing or sparse data.
Together, these pre-processing methods strengthen the training pipeline and ensure that TD-FusionUNet can effectively capture the complex patterns that underlie the spread of wildfires.

The proposed TD-FusionUNet can be implemented at the edge due to its lightweight architecture and reduced parameter count.
In addition, the TD layer is highly scalable and can be seamlessly integrated into other neural network architectures beyond U-Net \cite{pan2022deep}.
Its modular design allows it to replace or augment conventional convolutional layers, providing efficient transform-domain feature learning with minimal computational overhead.
\vspace{-4mm}
\section{TD-UNet Framework}
\label{sec:format}
\vspace{-5mm}
\subsection{Overview}
\vspace{-2mm}
The architecture of a single transform Unet is presented in Fig. \ref{fig1}, which is similar to Zhu's earlier network \cite{zhu2024probabilistichadamardunetmri}. The architecture comprises six convolutional blocks, each consisting of a 2D convolutional layer, a batch normalization layer, and a ReLU activation. Three of these blocks are in the encoder path, one is in the bottleneck, and the remaining two are used in the decoder. Between convolutional blocks, HT perceptron layers are inserted. Each of them applies a Hadamard Transform sequentially along 2D spatial dimensions, followed by element-wise multiplication with a trainable scaling weight matrix. The transformed features are then passed through a soft-thresholding module before being reconstructed via a 2D inverse Hadamard Transform \cite{agaian2006hadamard}, \cite{sarukhanyan2007reversible},\cite{zhu2024probabilistichadamardunetmri}.
Specifically, skip connections are formed by fusing selected convolutional and thresholding block outputs at different depths. These connections integrate features across HT/IHT stages in the transform domain, enhancing representational capacity without introducing additional heavy convolutional layers.

The multi-transform network is shown in Fig. \ref{DCTHT}. In the upper branch we use the DCT and in the lower part of the network we use the Hadamard Transform (HT). The outputs of these two branches are fused using a fully connected fusion layer as shown in Fig. \ref{DCTHT}.

In the following subsection we will go over the details of the HT-perceptron blocks. The DCT-perceptron blocks have the same structure except that the Hadamard transform is replaced by the well-known DCT.
\vspace{-4mm}
\subsection{HT-Layers: HT-Perceptron blocks}
\label{sec:print}
\vspace{-2mm}
The input tensor is represented as $\mathbf{X} \in \mathbb{R}^{C\times N\times N}$ where $C$ denotes the number of channels and $N$ denotes the image height and width, which are required to be powers of two ($N=64$ in our used dataset). The HT block maps each channel of the input represented by $\mathbf{X}_c$ to the Hadamard domain using the unnormalized Hadamard matrix $\mathbf{H}_N$ which satisfies $\mathbf{H}_N\mathbf{H}_N^\top = N\,\mathbf{I}$. As a result, the per-channel 2D transform is given by (\ref{eq:ht2d}).

\begin{equation}
\label{eq:ht2d}
\widehat{\mathbf{X}}_c = \mathbf{H}_N\,\mathbf{X}_c\,\mathbf{H}_N, \quad c=1,\dots,C.
\end{equation}

This operation achieves the 2D Hadamard Transform by applying the 1D Hadamard transform sequentially along both spatial dimensions: first along the rows by computing $\mathbf{H}_N \mathbf{X}_c$, and then along the columns by multiplying the result on the right with $\mathbf{H}_N$. 

Then the trainable weight block applies a weight map $\mathbf{W}_c\in\mathbb{R}^{N\times N}$ as entrywise scaling coefficients to the output of the HT block, as shown in (\ref{eq:scaling_index}).
\begin{equation}
\label{eq:scaling_index}
E_c(i,j) = W_c(i,j)\,\widehat{X}_c(i,j),\quad i,j=1,\dots,N.
\end{equation}
where $i$ and $j$ are indices of the spatial coordinates of the transform-domain feature map.

As a result, each Hadamard stage includes a transform-scale block for every channel. It is largely deterministic using Hadamard transform matrix multiplication with only lightweight learnable scaling, which reduces the need for numerous convolutional kernels and lowers parameter count.

To further promote efficiency, the soft-thresholding operator with nonnegative learnable thresholds is applied. The operator $S_T:\,\mathbb{R}\to\mathbb{R}$ is defined by (\ref{eq:soft_scalar}).  

\begin{equation}
\label{eq:soft_scalar}
S_T(e) = \begin{cases}
\operatorname{sgn}(e)\,(|e|-T) & |e|>T\\[4pt]
0 & |e|\le T
\end{cases}
\end{equation}
where $e$ represents $E_c(i,j)$, denoting the coefficient at spatial position $(i,j)$ in the scaled Hadamard spectrum, and $T$ is the corresponding nonnegative learnable threshold $T_c(i,j)$ with $\mathbf{T}_c \in \mathbb{R}_{\ge 0}^{N \times N}$.
The transform-domain activation is obtained component-wise as in (\ref{eq:soft_threshold_index}).
\begin{equation}
\label{eq:soft_threshold_index}
Z_c(i,j) = S_{T_c(i,j)}\big(E_c(i,j)\big)
\end{equation}
This yields the thresholded spectrum $\mathbf{Z}_c$. In this way, soft-thresholding further enhances efficiency by suppressing small coefficients to zero and allows only strong coefficients to pass through, forcing the network to focus on the most essential patterns. 

Finally, the feature map is reconstructed using the inverse Hadamard Transform (IHT) associated with the unnormalized forward HT, as shown in (\ref{eq:iht2d}), similar to the forward transform.
\begin{equation}
\label{eq:iht2d}
\mathbf{Y}_c = \frac{1}{N^2}\mathbf{H}_N\mathbf{Z}_c\mathbf{H}_N\quad c=1,\dots,C.
\end{equation}

\begin{table*}[!t]
\centering
\caption{Comparison of CNN Autoencoder, HT-UNet and TD-FusionNet in Google dataset \cite{huot2022wildfire}: Fire Spread Prediction}
\label{tab:cnn_htunet_comparison}
\renewcommand{\arraystretch}{1} 
\setlength{\tabcolsep}{9pt}     
\begin{tabular}{|l|c|l|l|c|c|c|c|}
\hline
\textbf{Model} & \textbf{Parameters} & \textbf{Preprocessing} & \textbf{Prediction} & \textbf{Precision} & \textbf{Recall} & \textbf{IoU} & \textbf{F1} \\ \hline
\multirow{2}{*}{2D-CNN Autoencoder} 
  & \multirow{2}{*}{78k} & Standard   & Next day  & 
  0.346 & 0.373 & 0.153 & 0.359 \\ \cline{3-8}
  &                         & Customized & Both days & 0.588 & 0.658 & 0.382 & 0.621 \\ \hline
\multirow{2}{*}{2D-HT-UNet}   
  & \multirow{2}{*}{45k} & Standard   & Next day  & 0.571 & 0.453 & 0.188$\uparrow$ & 0.505$\uparrow$ \\ \cline{3-8}
  &                         & Customized & Both days & 0.737 & 0.606 & 0.433$\uparrow$ & 0.665$\uparrow$ \\ \hline

TD-FusionUNet (base=8)
  &93k  & Customized & Both days 
  & 0.876 & 0.586 & 0.555 $\uparrow$ & 0.702$\uparrow$ \\ \hline
\end{tabular}
\end{table*}


\begin{table*}[t]
\centering
\caption{Comparison of baseline models and the proposed TD-FusionUNet on the WildfireSpreadTS dataset \cite{gerard2023wildfirespreadts}}
\label{tab:wildfire_DCT-HT-FusionUNet}
\renewcommand{\arraystretch}{1.15}
\setlength{\tabcolsep}{5pt}

\begin{tabular}{|l|c|c|c|c|c|c|}
\hline
\textbf{Model} & \textbf{Parameters}  & \textbf{Prediction} & \textbf{Precision} & \textbf{Recall} & \textbf{F1} \\
\hline
Baseline UNet (ResNet18) 
& 14M  & Next day 
& 0.536 & 0.654 & 0.589 \\
\hline
HT-UNet 
& 169k & Next day 
& 0.510 & 0.646 & 0.570 \\
\hline
TD-FusionUNet (base=4) 
& 159k  & Next day 
& 0.580 & 0.595 & 0.588 \\
\hline
\textbf{TD-FusionUNet (base=8)} 
& \textbf{370k} & \textbf{Next day} 
& \textbf{0.539} & \textbf{0.653} & \textbf{0.591$\uparrow$} \\
\hline
\end{tabular}
\vspace{-4mm}
\end{table*}
\subsection{Dual-Branch Network Extension}
\vspace{-2mm}
To examine the scalability of the proposed TD-FusionUNet architecture, we introduce a capacity-scaled dual-branch extension termed TD-FusionUNet, as illustrated in Fig.\ref{DCTHT}. In this way, we double the number of coefficients of the single branch HT-UNet. To further increase the number of parameters one can add another orthogonal transform branch etc.

Given an input tensor $\mathbf{X}\in\mathbb{R}^{C\times N\times N}$, two parallel encoding branches are constructed. The first branch follows the HT-based encoder described in the previous section, while the second branch applies a channel-wise 2D Discrete Cosine Transform (DCT).

Let the orthonormal DCT-II basis matrix be $\mathbf{D}_N$. The transform-domain latent space parameters are computed as follows
\begin{equation}
\widetilde{\mathbf{X}}_c = \mathbf{D}_N\,\mathbf{X}_c\,\mathbf{D}_N^\top,\quad c=1,\dots,C.
\end{equation}

Both branches follow an identical encoder--decoder hierarchy, where transform-domain features are processed by lightweight convolutional blocks and progressively downsampled in the encoder.

For the DCT-based branch, decoder reconstruction is performed using the inverse discrete cosine transform, where the spatial-domain features are recovered as $\mathbf{Y}_c = \mathbf{D}_N^\top \mathbf{Z}_c \mathbf{D}_N$ for each channel $c = 1,\dots,C$, with $\mathbf{Z}_c$ denoting the processed transform-domain coefficients after convolution.

To increase representational capacity in a controlled manner, the base channel width is scaled to 8, resulting in approximately 370k parameters.
Let $\mathbf{F}^{\mathrm{HT}}$ and $\mathbf{F}^{\mathrm{DCT}}$ denote the feature maps at the outputs of the HT and DCT branches, respectively.
During decoding, features are bilinearly upsampled, channel-aligned by convolution, and fused via element-wise summation:
\begin{equation}
\mathbf{F} = \phi(\mathbf{F}^{\mathrm{HT}}) + \psi(\mathbf{F}^{\mathrm{DCT}}),
\end{equation}
where $\phi(\cdot)$ and $\psi(\cdot)$ denote learnable convolutional mappings as shown in Fig. \ref{DCTHT} .

This design preserves the transform-domain inductive bias of the HT-based backbone while enabling complementary interaction between Hadamard-domain global mixing and DCT-domain frequency representations across the decoding hierarchy.

\noindent
{\em Customized data preprocessing}:\\
To mitigate the sparsity of fire pixels and enhance generalization, we introduce two preprocessing strategies tailored for sparse maps with missing data and binary masks.

\textit{Random margin cropping}. First, uncertain pixels ($-1$) are set to zero. A random margin adjustment is then applied to perturb the binary mask values. Specifically, background pixels ($0$) are replaced with small random values sampled uniformly as
$
x \sim \mathcal{U}(0.01, 0.03),
$
while fire pixels ($1$) are replaced with values sampled from
$
x \sim \mathcal{U}(0.8, 0.99).$

This procedure enriches the representation of both burned and unburned classes and prevents the network from overfitting to strictly binary masks.

\textit{Gaussian-Mixture Data Smoothing}. This procedure using a predefined set of standard deviations $\{\sigma_k\}$ (e.g., $\{0.4, 0.8\}$ used in our experiment). 
For each $\sigma_k$, a Gaussian blur is applied to the input mask, producing a softened map
\begin{equation}
\mathbf{G}_{\sigma_k} = \operatorname{GaussBlur}(\mathbf{M}, \sigma_k),
\end{equation}
where $\mathbf{M}$ denotes the input binary mask or other important input features with missing data.  

The resulting blurred maps $\{\mathbf{G}_{\sigma_k}\}$ are aggregated across scales 
using a mean operator to form the final soft probability map
\begin{equation}
\mathbf{P} = \frac{1}{K}\sum_{k=1}^K \mathbf{G}_{\sigma_k},
\end{equation}
where $K$ is the number of Gaussian scales.  
\vspace{-0.5 cm}
\section{Experimental Results}
\label{sec:pagestyle}
\vspace{-0.2 cm}
We use two publicly available datasets. The first dataset was released by Google Research \cite{huot2022wildfire}, which provides a comprehensive set of environmental and contextual features derived from satellite imagery and weather data. 
This dataset is split into training, validation, and test sets with an 8:1:1 ratio.

The single-branch Hadamard Transform-UNet (HT-UNet) model was implemented in PyTorch, using Spyder as the development environment and Anaconda for environment management. Since the existing dataset pipeline and baseline CNN autoencoder were implemented in TensorFlow, customized preprocessing routines were also developed in TensorFlow to ensure compatibility. 

The input images are $64 \times 64$ with 12 channels. The first convolutional layer in the encoder and the final transposed convolutional layer in the decoder use $4 \times 4$ filters, while all intermediate convolutional layers employ $7 \times 7$ filters. The HT-UNet outputs a probability map, and a threshold of 0.5 is applied to generate the final binary mask.

The training loss is defined as a weighted combination of three terms:
\begin{equation}
\mathcal{L} = \lambda_{\text{BCE}} \, \mathcal{L}_{\text{BCE}}
            + \lambda_{\text{Dice}} \, \mathcal{L}_{\text{Dice}}
            + \lambda_{\text{Focal}} \, \mathcal{L}_{\text{Focal}},
\end{equation}
where $\lambda_{\text{BCE}}=0.4$, $\lambda_{\text{Dice}}=0.3$, and $\lambda_{\text{Focal}}=0.3$. Here, $\mathcal{L}_{\text{BCE}}$ is the weighted binary cross-entropy loss, 
$\mathcal{L}_{\text{Dice}}$ is the Dice loss, and 
$\mathcal{L}_{\text{Focal}}$ is the focal loss. 

Training is performed for 100 epochs using the Adam optimization algorithm with a learning rate of $1 \times 10^{-4}$.
The performance metrics include the Intersection over Union (IoU), Precision, Recall, and the F1-score. The proposed HT-UNet achieves better results than the baseline 2D-CNN autoencoder network used in \cite{huot2022wildfire}, while reducing the number of parameters by almost 50\%, as shown in Table \ref{tab:cnn_htunet_comparison}, 
  which compares the baseline CNN Autoencoder and the proposed HT-UNet under two preprocessing settings.
  In the standard setting, the models are trained and evaluated on the original dataset to predict the next-day fire mask only. In the customized setting, random cropping was applied to the pre-fire mask, and the mixed Gaussian softening model was applied to both the pre-fire mask and wind speed, as these were considered the two most critical features \cite{shadrin2024wildfire}. The predefined deviations in the mixed Gaussian softening model are selected as $\{0.4,\,0.8\}$ based on multiple trials. Beyond these custom methods, the preprocessing pipeline also included standard operations such as shuffling, flipping, and normalization. 

The results show that HT-UNet consistently outperforms the CNN autoencoder in IoU and F1-score under both settings. Under the standard preprocessing, HT-UNet improves the IoU from 0.153 to 0.188 and the F1-score from 0.359 to 0.505. Both models benefit from the preprocessing enhancements, but HT-UNet achieves better performance, reaching an IoU of 0.433 and F1-score of 0.665, demonstrating its superior ability to capture wildfire spread patterns.  TD-FusionUNet has slightly more parameters than both HD-UNet and Convolutional autoencoder but it produces significantly better results than the other two networks, reaching an F1-score of 0.702 (the last row of Table 1).

In addition, we evaluated the single-branch HT-UNet and TD-FusionUNet on the WildFireSpreadTS dataset \cite{gerard2023wildfirespreadts}. 
The input images are $128 \times 128$ with 40 channels. 
The models were optimized using Adam with a learning rate of $1 \times 10^{-4}$ and trained for 300 epochs. We studied TD-FusionUNet with channel widths of 4, 8, and 16. The best performance is obtained when the channel width is 8, as shown in Table \ref{tab:wildfire_DCT-HT-FusionUNet},
which compares the proposed TD-FusionUNet with the ResNet18-based UNet baseline used in \cite{gerard2023wildfirespreadts}, and lightweight HT-based variants. 
The TD-FusionUNet with a base channel width of 8 achieves the best F1-score of 0.591 when trained with the composite loss (10), while using only 370k parameters, which is substantially fewer than the 14M parameters required by the ResNet18-based UNet  \cite{gerard2023wildfirespreadts}, demonstrating a favorable accuracy--efficiency trade-off. The inclusion of the base-4 TD-FusionUNet further indicates that the observed improvement is not solely due to increased parameter count, but arises consistently from the proposed transform-domain design as model capacity scales. 

\begin{figure}[htb]
\centering

\includegraphics[width=8.5cm]{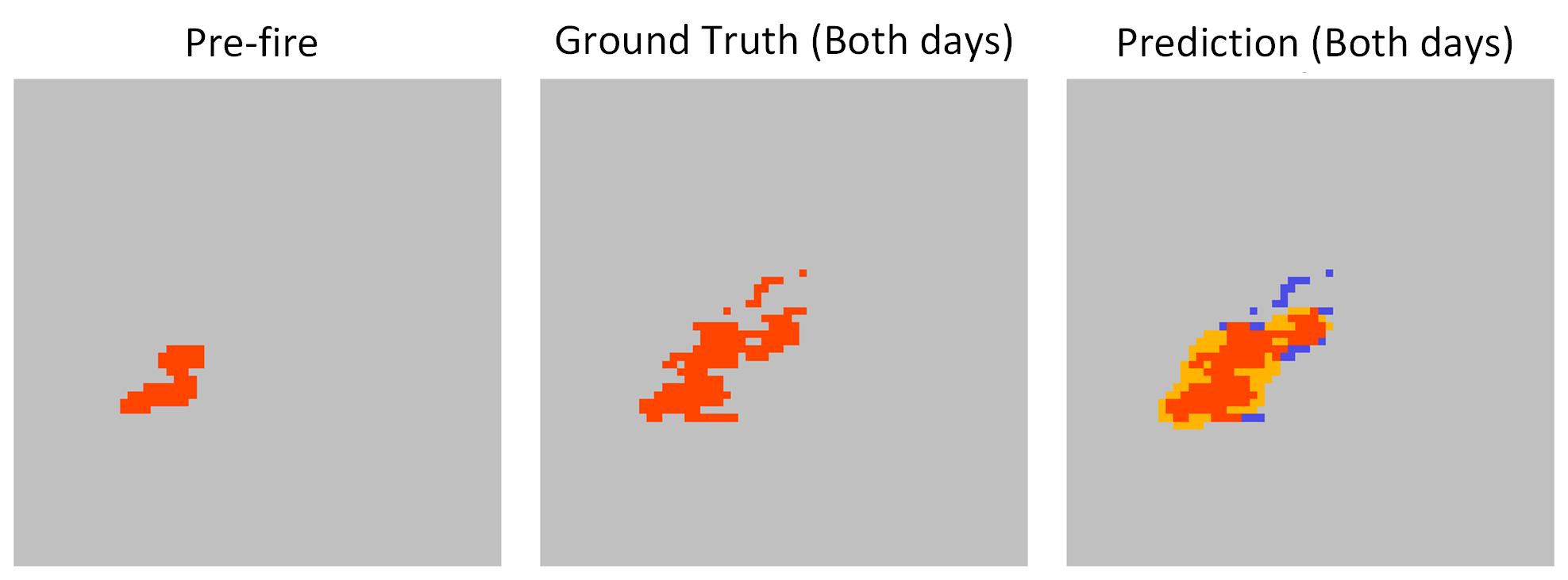}

\centerline{(a) Prediction Result 1}

\medskip
\includegraphics[width=8.5cm]{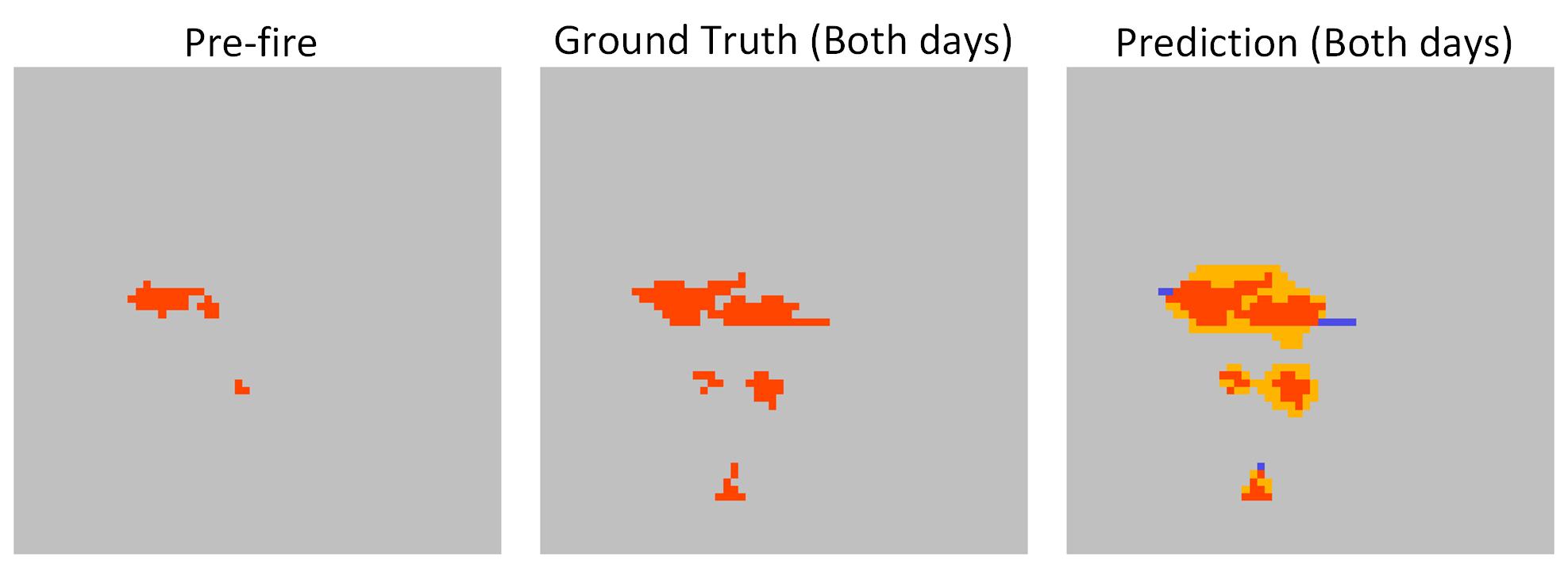}

\centerline{(b) Prediction Result 2}

\medskip
\includegraphics[width=8.5cm]{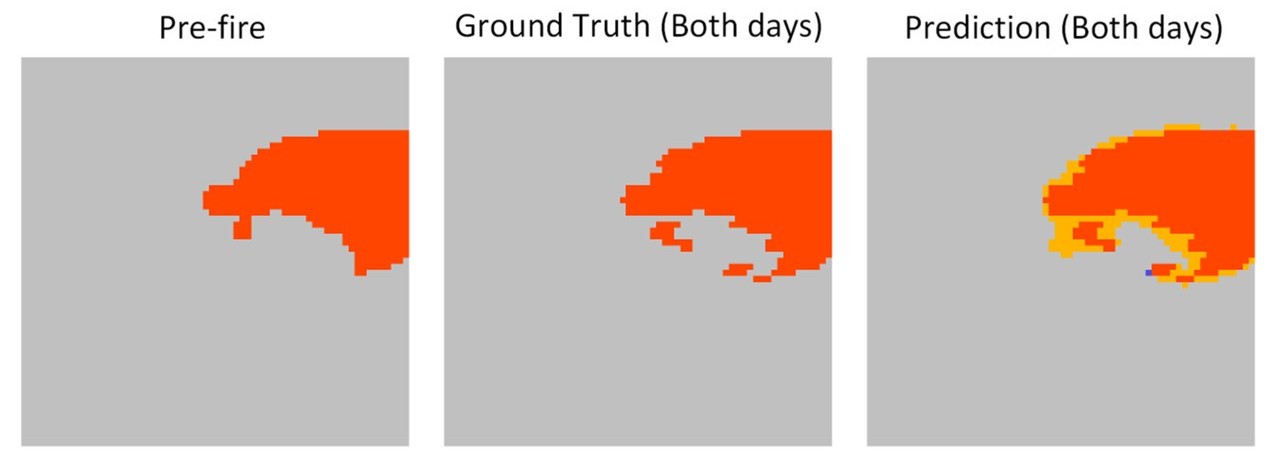}

\centerline{(c) Prediction Result 3}

\medskip
\caption{Examples of inference results, where false positives (FP) are highlighted in yellow and false negatives (FN) are highlighted in blue.}
\label{fig2}
\end{figure}

Fig. \ref{fig2} (a) and (b) show the prediction binary masks generated by the HT-UNet for both-day prediction in Google dataset \cite{huot2022wildfire}. It can be seen from both prediction examples that the predicted fire regions cover most of the ground truth areas (orange). In Fig. \ref{fig2} (c) a prediction result from WildFireSpreadTS dataset \cite{gerard2023wildfirespreadts} using the TD-FusionUNet is shown.  These examples demonstrate that the proposed models are capable of accurately capturing the spatial extent of fire spread, with relatively few false negatives (missed fire regions) and manageable false positives (over-predicted areas). 

\section{ Conclusions}
We developed the HT-UNet and TD-FusionUNet, which enable transform-domain learning with significantly fewer parameters than a typical UNet. Orthogonal Transform-domain methods learn better than regular UNets when the training dataset is small, thanks to their inherent procedure that forms an orthogonal latent space. They outperform the regular autoencoders and UNets in wild-fire spread prediction problem. 
Building upon the HT-UNet backbone, this paper introduces a TD-FusionUNet and evaluates its performance on the WildfireSpreadTS dataset. While maintaining a lightweight model size of only 370K parameters, the proposed fusion model with a base channel width of 8 achieves the best F1-score of 0.591, outperforming 
the ResNet18-based UNet baseline.


\bibliographystyle{IEEEbib}
\bibliography{main}

@techreport{nicc2024,
  title        = {Wildland Fire Summary and Statistics Annual Report 2024},
  author       = {{National Interagency Coordination Center}},
  year         = {2024},
  institution  = {U.S. Department of the Interior, Boise National Interagency Fire Center},
  note         = {Retrieved 2025-09-20}
}

@article{gunay2012entropy,
  title={Entropy-functional-based online adaptive decision fusion framework with application to wildfire detection in video},
  author={Gunay, O. and Toreyin, B. U. and Kose, K. and Cetin, A. E.},
  journal={IEEE Transactions on Image Processing},
  volume={21},
  number={5},
  pages={2853--2865},
  year={2012},
  publisher={IEEE}
}

@inproceedings{pan2022deep,
  title={Deep neural network with Walsh-Hadamard transform layer for ember detection during a wildfire},
  author={Pan, H. and Badawi, D. and Chen, C. and Watts, A. and Koyuncu, E. and Cetin, A. E.},
  booktitle={Proceedings of the IEEE/CVF CVPR},
  pages={257--266},
  year={2022}
}

@misc{calfire2025incidents,
  title        = {{CALFIRE: Current Emergency Incidents}},
  author       = {{California Department of Forestry and Fire Protection (CAL FIRE)}},
  year         = {2025},
  howpublished = {\url{https://www.fire.ca.gov/incidents}},
  note         = {Archived from the original on Jan. 30, 2025. Retrieved Feb. 2, 2025}
}

@article{iglesias2022fires,
  title        = {U.S. fires became larger, more frequent, and more widespread in the 2000s},
  author       = {Iglesias, V. and Balch, J. K. and Travis, W. R.},
  journal      = {Science Advances},
  year         = {2022},
  publisher    = {AAAS},
  volume       = {8},
  number       = {12},
  doi          = {10.1126/sciadv.abm2219}
}

@incollection{ostoja2023nca,
  author       = {Ostoja, S. M. and Crimmins, A. R. and Byron, R. G. and East, A. E. and Méndez, M. and O’Neill, S. M. and Peterson, D. L. and Pierce, J. R. and Raymond, C. and Tripati, A. and Vaidyanathan, A.},
  title        = {Focus on Western Wildfires},
  booktitle    = {Fifth National Climate Assessment},
  publisher    = {USGCRP},
  year         = {2023},
  doi          = {10.7930/NCA5.2023.F2},
  url          = {https://doi.org/10.7930/NCA5.2023.F2}
}

@article{singh2025review,
  title={A Comprehensive Review of Empirical and Dynamic Wildfire Simulators and Machine Learning Techniques used for the Prediction of Wildfire in Australia},
  author={Singh, Harikesh and Ang, Li-Minn and Paudyal, Dipak and Acuna, Mauricio and Srivastava, Prashant Kumar and Srivastava, Sanjeev Kumar},
  journal={Technology, Knowledge and Learning},
  pages={1--34},
  year={2025},
  publisher={Springer}
}

@article{singh2021land,
  author       = {Singh, H. and Pandey, A. C.},
  title        = {Land Deformation Monitoring Using Optical Remote Sensing and {PSInSAR} Technique Nearby Gangotri Glacier in Higher Himalayas},
  journal      = {Modeling Earth Systems and Environment},
  volume       = {7},
  number       = {1},
  pages        = {221--233},
  year         = {2021},
  doi          = {10.1007/s40808-020-00889-5},
  publisher    = {Springer}
}

@article{duff2021wildfire,
  author       = {Duff, T. J. and Penman, T. D.},
  title        = {Determining the Likelihood of Asset Destruction During Wildfires: Modelling House Destruction with Fire Simulator Outputs and Local-Scale Landscape Properties},
  journal      = {Safety Science},
  volume       = {139},
  pages        = {105196},
  year         = {2021},
  doi          = {10.1016/j.ssci.2021.105196},
  publisher    = {Elsevier}
}

@article{eden2020fire,
  author       = {Eden, J. M. and Krikken, F. and Drobyshev, I.},
  title        = {An Empirical Prediction Approach for Seasonal Fire Risk in the Boreal Forests},
  journal      = {International Journal of Climatology},
  volume       = {40},
  number       = {5},
  pages        = {2732--2744},
  year         = {2020},
  doi          = {10.1002/joc.6363},
  publisher    = {Wiley}
}

@incollection{rahman2018forest,
  author       = {Rahman, S. and Chang, H.-C. and Magill, C. and Tomkins, K. and Hehir, W.},
  title        = {Forest Fire Occurrence and Modeling in Southeastern Australia},
  booktitle    = {Forest Fire},
  publisher    = {InTech},
  year         = {2018},
  doi          = {10.5772/intechopen.76072},
  url          = {https://doi.org/10.5772/intechopen.76072}
}

@article{digiuseppe2025fire,
  author       = {Di Giuseppe, F. and McNorton, J. and Lombardi, A. and others},
  title        = {Global Data-Driven Prediction of Fire Activity},
  journal      = {Nature Communications},
  volume       = {16},
  pages        = {2918},
  year         = {2025},
  doi          = {10.1038/s41467-025-58097-7},
  publisher    = {Nature Publishing Group}
}

@article{hodges2019wildland,
  author       = {Hodges, J. L. and Lattimer, B. Y.},
  title        = {Wildland Fire Spread Modeling Using Convolutional Neural Networks},
  journal      = {Fire Technology},
  volume       = {55},
  pages        = {2115--2142},
  year         = {2019},
  doi          = {10.1007/s10694-019-00846-4},
  publisher    = {Springer}
}

@article{kantarcioglu2023fire,
  author       = {Kantarcioglu, O. and Kocaman, S. and Schindler, K.},
  title        = {Artificial Neural Networks for Assessing Forest Fire Susceptibility in {T{\"u}rkiye}},
  journal      = {Ecological Informatics},
  volume       = {75},
  pages        = {102034},
  year         = {2023},
  issn         = {1574-9541},
  doi          = {10.1016/j.ecoinf.2023.102034},
  publisher    = {Elsevier}
}

@inproceedings{Radke2019FireCastLD,
  title={FireCast: Leveraging Deep Learning to Predict Wildfire Spread},
  author={Radke, D. and Hessler, A. and Ellsworth, D.},
  booktitle={Int. Joint Conf. Artificial Intelligence},
  year={2019},
  url={https://api.semanticscholar.org/CorpusID:199466194}
}

@article{huot2022wildfire,
  title={Next day wildfire spread: A machine learning dataset to predict wildfire spreading from remote-sensing data},
  author={Huot, Fantine and Hu, R Lily and Goyal, Nita and Sankar, Tharun and Ihme, Matthias and Chen, Yi-Fan},
  journal={IEEE Transactions on Geoscience and Remote Sensing},
  volume={60},
  pages={1--13},
  year={2022},
  publisher={IEEE}
}

@article{shadrin2024wildfire,
  author       = {Shadrin, D. and Illarionova, S. and Gubanov, F. and others},
  title        = {Wildfire Spreading Prediction Using Multimodal Data and Deep Neural Network Approach},
  journal      = {Scientific Reports},
  volume       = {14},
  pages        = {2606},
  year         = {2024},
  doi          = {10.1038/s41598-024-52821-x},
  publisher    = {Nature Publishing Group}
}

@misc{dong2017learningdeeprepresentationsusing,
  title  = {Learning Deep Representations Using Convolutional Auto-encoders with Symmetric Skip Connections},
  author = {Dong, J. and Mao, X.-J. and Shen, C. and Yang, Y.-B.},
  year   = {2017},
  note   = {arXiv:1611.09119}
}

@misc{ronneberger2015unetconvolutionalnetworksbiomedical,
  title  = {U-Net: Convolutional Networks for Biomedical Image Segmentation},
  author = {Ronneberger, O. and Fischer, P. and Brox, T.},
  year   = {2015},
  note   = {arXiv:1505.04597}
}

@misc{zhou2018unetnestedunetarchitecture,
  title  = {UNet++: A Nested U-Net Architecture for Medical Image Segmentation},
  author = {Zhou, Z. and Siddiquee, M. M. R. and Tajbakhsh, N. and Liang, J.},
  year   = {2018},
  note   = {arXiv:1807.10165}
}

@book{agaian2006hadamard,
  title={Hadamard matrices and their applications},
  author={Agaian, S. S.},
  volume={1168},
  year={2006},
  publisher={Springer}
}

@article{sarukhanyan2007reversible,
  title   = {Reversible Hadamard Transforms},
  author  = {Sarukhanyan, Hovhannes and Agaian, Sos and Egiazarian, Karen and Astola, Jaakko},
  journal = {Facta Universitatis, Series: Electronics and Energetics},
  volume  = {20},
  number  = {3},
  pages   = {309--330},
  year    = {2007}
}

@article{zhu2024probabilistichadamardunetmri,
  title  = {A Probabilistic Hadamard U-Net for MRI Bias Field Correction},
  author = {Zhu, X. and Pan, H. and Velichko, Y. and Murphy, A. B. and Ross, A. and Turkbey, B. and Cetin, A. E. and Bagci, U.},
  year   = {2025},
  journal={accepted for publication, Medical Image Analysis},
   note   = {arXiv:2403.05024}
}

@article{gerard2023wildfirespreadts,
  title={Wildfirespreadts: A dataset of multi-modal time series for wildfire spread prediction},
  author={Gerard, Sebastian and Zhao, Yu and Sullivan, Josephine},
  journal={Advances in Neural Information Processing Systems},
  volume={36},
  pages={74515--74529},
  year={2023}
}

\end{document}